\documentclass[letterpaper, 10 pt, conference]{ieeeconf}  

\IEEEoverridecommandlockouts                              

\usepackage{times}

\usepackage{multicol}
\usepackage[bookmarks=true]{hyperref}
\usepackage{graphicx,import} 
\usepackage{transparent}
\usepackage{amsmath} 
\usepackage{amssymb}  
\usepackage{commath}
\usepackage{paralist}
\usepackage{algorithm}
\usepackage{algpseudocode}
\algnewcommand{\IfThenElse}[3]{
  \State \algorithmicif\ #1\ \algorithmicthen\ #2\ \algorithmicelse\ #3}
\algnewcommand\algorithmicforeach{\textbf{for each}}
\algdef{S}[FOR]{ForEach}[1]{\algorithmicforeach\ #1\ \algorithmicdo}
\usepackage{lipsum}  
\usepackage{epstopdf}
\usepackage{filecontents}
\usepackage{color}
\usepackage{xcolor}
\usepackage{xxcolor}
\usepackage{pgf}
\usepackage{pgfplots}
\usepackage{tikz}
\usepackage{setspace}
\usepackage[skins]{tcolorbox}
\usepackage{todonotes}
\usepackage{multirow}
\usepackage{array}
\usepackage{svg}
\usepackage{footmisc}
\usepackage{url}

\title{\LARGE \bf
Hierarchical Control of Smart Particle Swarms
}

\author{Vivek Shankar Varadharajan, Sepand Dyanatkar and Giovanni Beltrame
\thanks{Varadharajan V. S. and Beltrame G. are with Polytechnique Montreal, 2500 Chem. de Polytechnique, Montreal, QC H3T 1J4. (email: vivek-shankar.varadharajan@polymtl.ca, giovanni.beltrame@polymtl.ca)}
\thanks{Dyanatkar S. is with The University of British Columbia, 2329 West Mall, Vancouver, BC V6T 1Z4. (email: sepandd@student.ubc.ca)}
}

\begin{document}

\maketitle
\thispagestyle{empty}
\pagestyle{empty}

\begin{abstract}
  We present a method for the control of robot swarms using two subsets of
  robots: a larger group of simple, oblivious robots (which we call \emph{the
    workers}) that is governed by simple local attraction forces, and a smaller
  group (the \emph{guides}) with sufficient mission knowledge to create and
  displace a desired worker formation by operating on the local forces of the
  workers. The guides coordinate to shape the workers like smart particles by
  changing their interaction parameters. We study the approach with a large
  scale experiment in a physics based simulator with up to 5000 robots forming
  three different patterns. Our experiments reveal that the approach scales well
  with increasing robot numbers, and presents little pattern distortion. We
  evaluate the approach on a physical swarm of robots that use visual inertial
  odometry to compute their relative positions and obtain results that are
  comparable with simulation. This work lays the foundation for designing and
  coordinating configurable smart particles, with applications in smart
  materials and nanomedicine.
\end{abstract}

\section{Introduction}
Decentralized robot swarms have many desirable properties like scalability, robustness,
and flexibility~\cite{Dorigo2021}.
Robot swarms are generally composed of a team of robots that are simple and
inexpensive, but when working together, they are capable of performing tasks
reserved to more complex and expensive robots. Some futuristic views on swarm
robotics~\cite{Dorigo2020} suggest that engineering methods for heterogeneous,
hierarchical self-organization is required in robot swarms to transition from
laboratories to real-world applications. This work is a step in this direction
that leverages self-organization through local interactions in a two-level robot
hierarchy. We posit that hierarchies in robot swarms can be of two types:
information hierarchies and intelligence hierarchies. Information hierarchies
consider identical robots with varying level of mission-related information: the
robots with more information can make better mission decisions and are placed at
the top of the hierarchy.
Intelligence hierarchies consider heterogeneous swarms where some robots have
capabilities that can lead to enhanced intelligence and decision-making, thus
placing them at top of the hierarchy.

Hierarchies allow for fewer complex robots with a global view of the mission,
while larger numbers of simpler robots perform the bulk of the work. As an
example, guides can lead nanorobots with limited sensing to deliver drugs
directly inside a cancer, where they would not be able to enter due to their
size.
We posit that hierarchies in robots enable building complex behaviors in a more
traditional way while retaining the properties of scalability, fault tolerance,
and cost-effectiveness of robot swarms.


\begin{figure}
    \centering
    \includegraphics[width=0.99\linewidth]{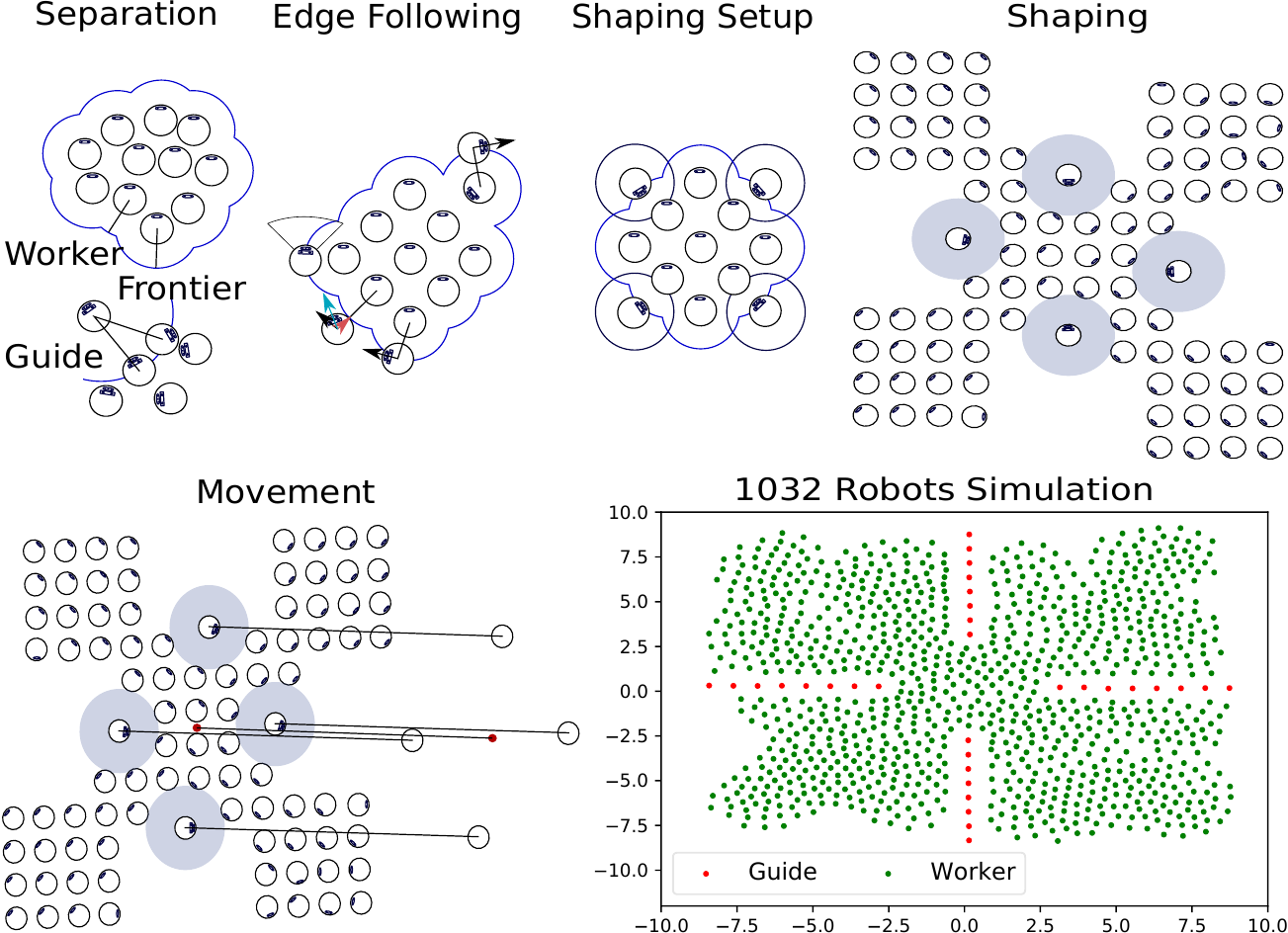}
    \caption{The states followed by the guide robots in our system.}
    \label{fig:guide_states}
\end{figure}

Hierarchical approaches to swarm intelligence have been investigated in the
context of particle swarm optimization~\cite{scheutz2007real,
  chen2010hierarchical}, but they have received very little attention in
robotics~\cite{Zhang2023}. The shepherding problem~\cite{hu2020occlusion,
  razali2013flock, ozdemir2017shepherding} considers a hierarchy with
\emph{sheep} that need to be herded by robot \emph{dogs}. The problem considered
in this work is comparable with shepherding, with the main difference being that
our \emph{guides} (or dogs in the shepherding analogy) shape and maintain the
pattern made by \emph{workers} (a.k.a. the sheep) by \textbf{operating on the
  rules that govern the workers swarm}.
Our approach also resembles smart particle
design~\cite{huang2015multifunctional} and finds potential applications in smart
materials and targeted drug delivery. Just like smart particles can be designed
to be dynamically programmed to change their attraction forces, a few
intelligent robot guides can order the workers to a desired shape and move them
to a target location by operating on the attractive and repulsive forces between
the workers. We open-source the code~\cite{particleswarmrepo} used in this work. 


\section{Related work}

Patterns in robot swarms are generally achieved by designing collective
behaviors that emerge as a result of local interactions with
neighbors~\cite{Slavkov2018}. Some examples of pattern formation are
morphogenesis~\cite{Slavkov2018} and self-assembly~\cite{Rubenstein795}. These
methods create stable structures using only local interactions.

\textbf{Flocking.} In some applications, robots are required to move while
maintaining a pattern or formation. This behavior is usually referred to as
flocking~\cite{Vasarhelyi2018, ramos2019evolving}, taking inspiration from the
behavior of birds. Flocking is extensively studied in
literature~\cite{fine2013unifying}, starting from the basic microscopic
model~\cite{reynolds1987flocks} of attraction, repulsion and cohesion.

\textbf{Leader-follower methods.} Some approaches use a leader-follower strategy
for flocking, where a few robots act as leaders~\cite{dimarogonas2006leader}
that share metrics and positions with the other robots, which in turn use this
information to coordinate their motion. The leader-follower problem is widely
studied~\cite{ji2006leader, panagou2014cooperative}. Centralized approaches use
techniques like algebraic graph theory~\cite{ji2006leader}, Hilbert space
projection, and dynamic programming~\cite{ji2006leader,bjorkenstam2006leader}.
Decentralized versions employ techniques like
tractor-trail~\cite{panagou2014cooperative} and virtual
springs~\cite{wiech2018virtual}.
Most of these methods are computationally intensive, with rare
exceptions~\cite{wiech2018virtual} that rely on local information and do not
suffer from scalability issues.
 
\textbf{Shepherding.} The leader-follower approach has its natural extension to
the shepherding problem: simple robots (the `sheep') that are herded by
one~\cite{hoshi2018robustness} or a group of robots (the
`shepherds')~\cite{chaimowicz2007aerial}. One common approach to shepherding is
to approximate a cluster of sheep as a circle or ellipse that needs to be
maintained over motion~\cite{chaimowicz2007aerial, razali2013flock,
  ozdemir2017shepherding} by placing the shepherds behind or around the sheep
herd.
Hu et. al.~\cite{hu2020occlusion} proposed an occlusion-based herding that places
the shepherds in the occluded region to the goal behind the sheep.
Other comparable approaches use Hilbert space filling and path
planning~\cite{long2021shepherding}, repulsive forces~\cite{kubo2022herd} and
connected-component labeling~\cite{razali2013flock}. Many approaches require a
reliable communication between the shepherd and sheep, besides a few that rely
on local measurements~\cite{ozdemir2017shepherding}. Similarly to flocking, most
shepherding approaches do not focus on maintaining a pattern during the
translation to a goal.

We consider a variation to the shepherding problem with two main differences:
first, we maintain the pattern during movement to the goal. Maintaining the
pattern is important for applications like nanomedicine or smart materials, as
the relative position of the particles might have important functional
implications; second, the shepherds (the guides in our parlance) have the
ability to change the interaction forces between the sheep (i.e., the workers),
which truly establishes a hierarchical control system on the swarm.

\begin{figure}[tbp]
    \centering
    \includegraphics[width=0.99\linewidth]{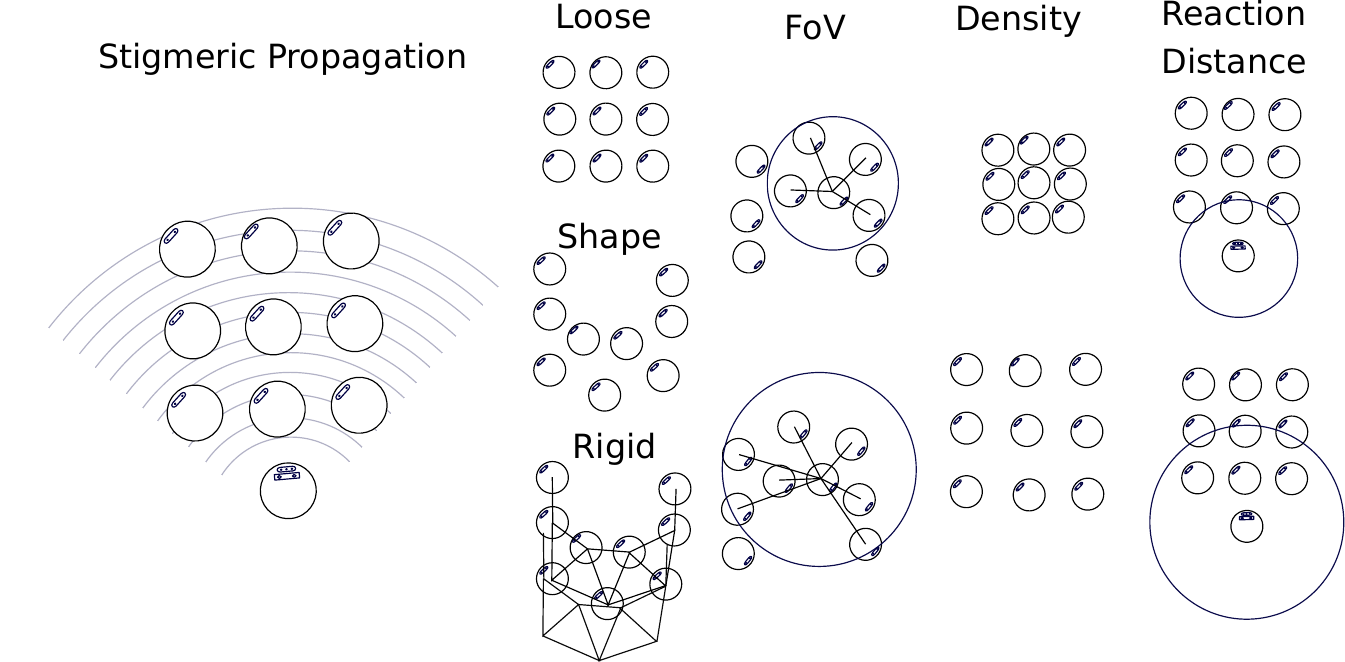}
    \caption{The control parameters of the workers,
      controlled by the guides.}
    \label{fig:worker_param}
\end{figure}

\section{Model and System}
Consider a group of $n$ robots with each robot $i$ having the following dynamics
\begin{equation}\label{equ:control_input}
  \dot{x}_{i} = v_{i}, v_{i} = u_{i}, i \in R, R=\{r_1, ..., r_n\}
\end{equation}
with $x_{i}$, $v_{i}$ and $u_{i}$ $\in \mathbb{R}^{2}$ being the position,
velocity and control input respectively. The relative position ($x_{ij}$) and
velocity ($v_{ij}$) of two robots $i$ and $j$ are $x_{ij}=x_{i} - x_{j}$ and
$v_{ij}=v_{i}-v_{j}$. We define the set of relative inter-agent distances $D =\{
d_{ij} = \norm{x_{i} - x_{j}}\in \mathbb{R} \vert i,j = 1,...,N, i \neq j\}$
and the communication graph formed between the robots as
$\mathcal{G} = (R,E,\textit{W})$ with $R = \{r_1, ..., r_n\}$ the set of vertices,
$E = \{ e_{ij} \vert i,j \in N, i \neq j \}$ the edge set and $W=\{w_{ij} \in R_+ |
e_{ij} \in E\}$ are weights assigned to the edges. Each robot $i \in R$ forms a
neighbors set $N_i =\{ j \in R \vert \norm{x_{ij}} \leq d_{com} \}$ as dictated
by $E$, with $d_{com}$ representing the communication range of the robots.

Guides and workers further divide the graph vertex set $R$ into $R^{G}$
(vertices corresponding to guides) and $R^{W}$ (vertices corresponding to
workers), with $R^{G} \cap R^{W} = \emptyset$. The guide robots are assumed to
be equipped with sophisticated sensors capable of performing localization, path
planning, traversability analysis, etc. The worker robots are simple and
oblivious, being capable of only communicating and measuring the relative
position $x_{ij}$ of their neighbors. The enhanced capabilities of the guide
robots place them in a leadership role: they perceive mission goals and provide
directives to the worker robots both indirectly (by moving to a different
position) and directly (by ordering a change to the attractive/repulsive forces
among the workers).

The neighbor set $N_i$ of robot $i$ can be divided into guide neighbors
$N^{G}_{i} = N_i \cup R^{G}$ and worker neighbors $N^{W}_{i} = N_i \cup R^{W}$.
The center of mass of the worker robots is defined as $x^{c} = \frac{1}{\lvert
  R^{W} \rvert }\sum_{i \in R^W } x_{i}$, and assumes a circular approximation of
the worker cluster. The robots are only capable of measuring relative positions
$x_{ij}$ and hence the center of mass perceived by robot $i$ is $x^{c}_{i} =
\frac{1}{\lvert N^{W}_{i} \rvert }\sum_{i \in N^{W}_{i} } x_{ij}$.

Fig.~\ref{fig:guide_states} illustrates the different states of the guide
robots. To shape a cluster of workers, the guides are given $k$ shaping targets,
each specified by an angle ($\theta^{s}$) and a distance ($d^s$) with respect to
the worker center of mass ($x^c$). Consider the shape angle set $\Theta^{s} = \{
\theta^{s}_1, ..., \theta^{s}_k \}$ and shape distance set $D^s= \{ d_1^s, ...,
d^s_k\}$. A guide robot $i \in R^{G}$ negotiates the lowest cost shaping target
$\theta_i \in \Theta^{s}$ (cost determined using distance to target) through a
decentralized task allocation mechanism from our previous
work~\cite{Varadharajan2020}. To reach its target angle $\angle{ (x_i - x^{c}) }
- \theta^s_i < \theta_\delta$, with $\theta_\delta$ being a small tolerance, a
guide $i$ follows the edge of the worker cluster. Upon reaching $\theta_i^s$,
the robots translate to ensure a distance $d_i^s$ from the center of mass of the
workers cluster, withing a small tolerance $d_\delta$: $d^s_i - \norm{x^{c}_{i}}
< d_\delta$. Once a shaping target is reached (both angle and distance), the
guides wait for all the other guides to reach their respective shape targets
using an agreement mechanism called a barrier~\cite{rosbuzz2020}.

Once at their shaping targets, the guides translate to a target distance
$d^{sp}$ from the center of mass of the worker cluster, thus forcing the workers
into the desired shape. We use a barrier to ensure all the guides have reached
their negotiated locations, upon which the guides start moving to a sequence of
targets $\mathbb{T} = \{x_1,...,x_n\}, \forall x_i \in \mathbb{R}^2$ maintaining
the worker formation. The guides interact with the workers through simple
potentials by changing their position, resulting in repulsive action on
the workers.

The guides can also manipulate the parameters of the potential fields that
control the workers. These parameters can be propagated across the swarm and
change the workers reactions to the actions of guides.
Fig.~\ref{fig:worker_param} summarizes the parameters: Density
($\rho$) defines the distance to maintain for robots in set $N^{W}_{i}$ for any
robot $i$ within the field of view ($FoV$). $\rho$ and $FoV$ influence the strength of
the robots in formation. The mode $M \in \{Loose,Shape,Rigid\}$ defines the
worker's behavior to maintain distance between neighbors. In \emph{Loose}, the
workers maintain a distance $\rho$ between all robots in $N^{W}_{i}$ with a
distance less than $Fov$. Workers store a tuple set $dM^{W}_{i} =\{ (j,d_{ij}) |
j \in N^{W}_{i}\}$ to keep a formation every time a mode switch to \emph{Shape}
or \emph{Rigid} happens. In \emph{Shape}, removal and addition of new robots to
$dM^{W}_{i}$ are allowed as they appear in the neighbors list $N^{W}_{i}$, in
contrast to \emph{Rigid} where $dM^{W}_{i}$ is frozen. In both \emph{Shape} and
\emph{Rigid}, the robots maintain the distances in $dM^{W}_{i}$ to preserve the
formation. The reaction distance $RFoV$ is the distance at which the workers
start responding to a guide robot.

\begin{figure}[tpb]
    \centering
    \includegraphics[width=0.99\linewidth]{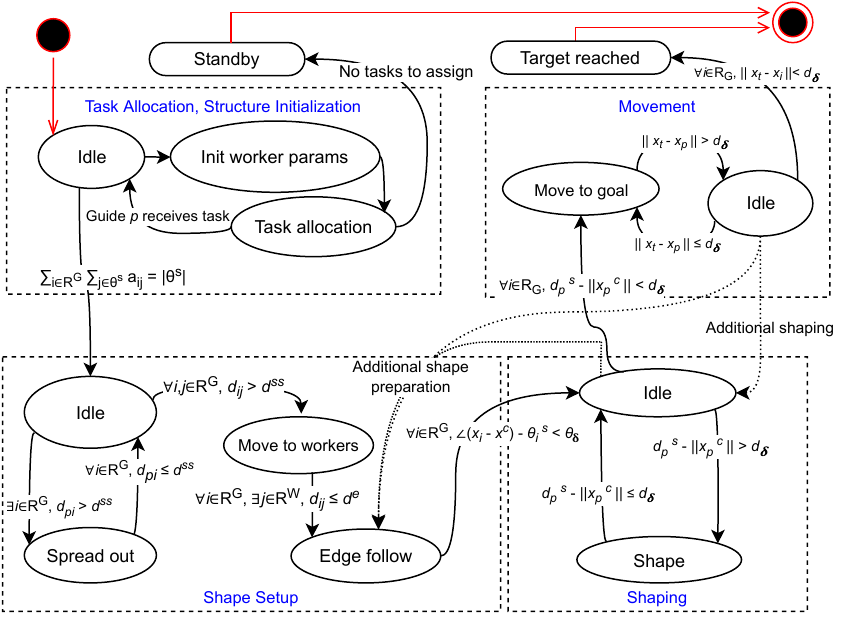}
    \caption{Finite state machine describing the high level guide behavior for
      all $i \in R^G$. The dotted boxes group the states into high-level states.
      The diagram shows the standard single shape objective edge following and
      shaping procedure in solid black connections. Multi-shape execution is shown with dotted connections. }
    \label{fig:guide_state_machine}
\end{figure}

\section{Methodology}

\begin{figure}
  \centering
  \includegraphics[width=0.85\linewidth]{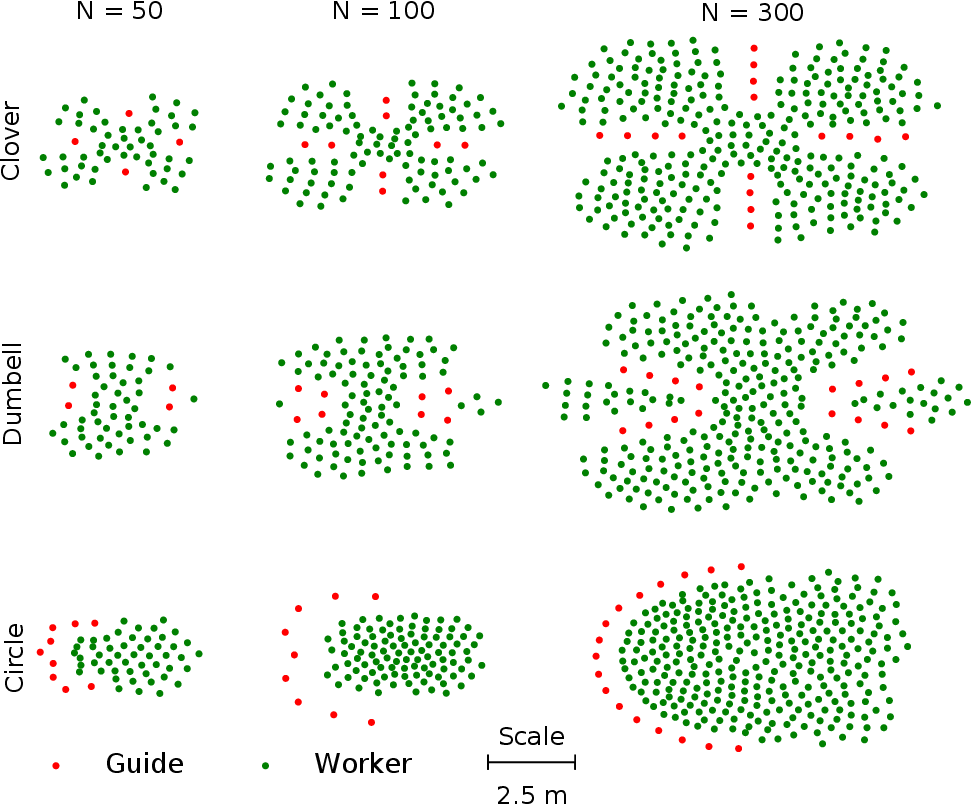}
  \caption{The range of shape configurations simulation used in the experiments, with varying swarm sizes. The snapshots are taken as the shaping mode is complete, before the guides transition to movement mode. }
  \label{fig:shapes}
\end{figure}

\subsection{Worker Swarm}

\begin{figure}[tbp]
  \centering
  \includegraphics[width=0.98\linewidth]{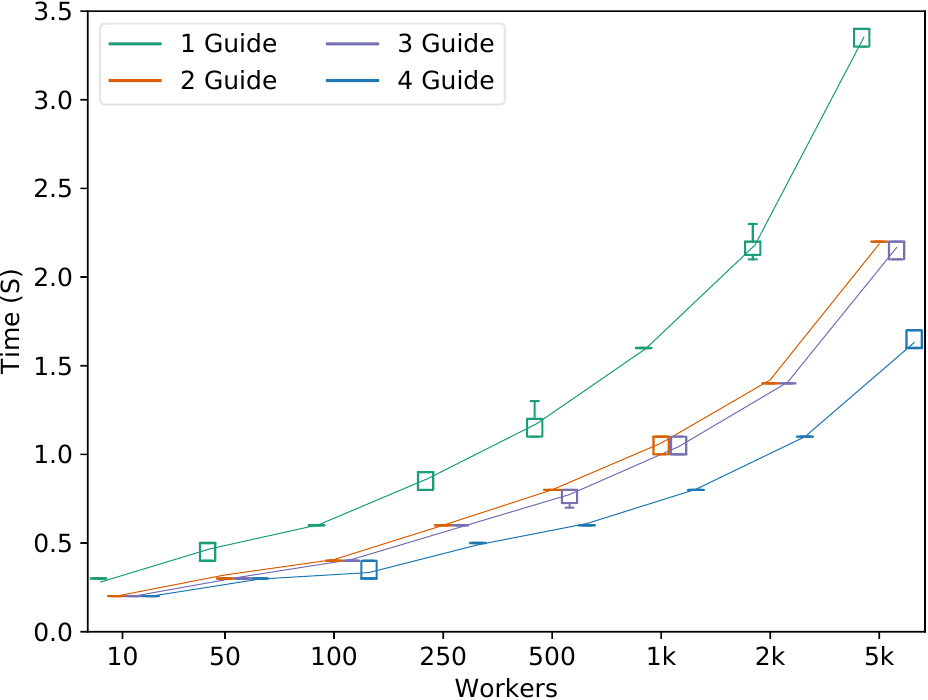}
  \caption{Time taken to propagate worker control parameters using different numbers of guide robots (Markers of different guide numbers are slightly offset for visual clarity).}
  \label{fig:convergence_cp}
\end{figure}

The four Control Parameters (CP)
of the worker robots are propagated by the guide robots using virtual
stigmergy~\cite{pinciroli2016tuple}, a conflict-free replicated data structure
that acts as a global distributed shared memory for the swarm. Virtual stigmergy
uses gossip-based communication that propagates through the network updating the
local copies of information on the robots, and was shown to converge if a
communication path exists (connected graph). Fig.~\ref{fig:convergence_cp} shows the convergence time of propagating a change in CP among the worker robots with various number of guides. The propagation of a change in CP takes about 1.5s with 4 guides for a 5k worker swarm and increases to under 3.5s with a single guide. 

A common method to design a control law for the sheep in
shepherding~\cite{hu2020occlusion} or flocking~\cite{Zhang2011} is to use
artificial potential functions that produce attraction and repulsion between the
robots. Unlike the shepherding problem, we need to maintain the desired
formation during its translation. Potential functions used in modeling the
bonding structure of atomic molecules~\cite{stuart2000reactive} are some
interesting candidates for enforcing a bond between two robots. We considered
various potentials like Morse (used in modeling diatomic molecules),
Lennard-Jones (used in modeling rare gases), Buckingham (used in modeling
ceramic materials) and the harmonic approximation (used in modeling crystalline
lattice structures)~\cite{stuart2000reactive}. These potentials produce uneven
attraction and repulsion forces when applied on robots, for instance produce
stronger repulsion and weaker attraction becomes a problem when these parameters
are dynamically configured by the guide robots. We therefore choose the
harmonic approximation to define the bonding force between the worker robots. A
standardized Harmonic potential is of the following form $\phi(x) = \frac{1}{2}
k (x)^2$, we empirically shape the potential to obtain a desired force response
for a given distance and obtain the potential function:
\begin{equation}
  \label{equ:harmonic_potential}
  \phi(d) = a_0 + \frac{k d^3}{2\lvert d \rvert}
\end{equation}
with $a_0$ and $k$ being design constants. The feedback control contribution for
maintaining the formation between other worker robots within the field of view (determined based on the current mode) is $w_{i}$ and the control contribution for producing a repulsive
reaction from the guide robots within the reactive field of view $RFoV$ is
$g_{i}$
\begin{align}
\label{equ:worker_control}
u_{i} &= \alpha w_{i} + \beta g_{i} \\
\label{equ:worker_control:contri1}
w_{i} &= \frac{1}{\lvert N^{FoV}_{i} \rvert} \sum_{j \in N^{FoV}_{i}} \phi(d_{ij} - d_{\rho}) \frac{x_{ij}}{\norm{x_{ij}}} \\
\label{equ:worker_control:contri2}
g_{i} &= \frac{1}{\lvert N^{RFoV}_{i} \rvert} \sum_{j \in N^{RFoV}_{i} } \phi^+(d_{ij} - d_{\rho}) \frac{x_{ij}}{\norm{x_{ij}}}
\end{align}

The potential function $\phi(d_{ij} - d_{\rho})$ provides the magnitude and 
$\frac{x_{ij}}{\norm{x_{ij}}}$ determines the direction of each $w_{ij}$ and $g_{ij}$ contribution. $\alpha$
and $\beta$ are design parameters that define the relative importance of
maintaining formation with respect to reacting to the guides in proximity. The
function $\phi^+(d) \to \mathbb{R}^+$ produces only repulsion mapping to
positive real values if and only if the input domain is $d \in {[} 0, \infty {]}$.
Similarly, the function $\phi(d) \to \mathbb{R}$ produces real values when 
$d \in {[} -\infty,\infty{]}$. 
Both functions $\phi^+(.)$ and $\phi(.)$ are
Lipschitz continuous and belong to class function $\phi^+(.) \in S^+$ and
$\phi(.) \in S$. The class functions are defined as $S^+=\{ f:{[}0,\infty {)} \to
\mathbb{R} \vert f(d) = 0, \forall d \leq 0, f(d) > 0, \forall d > 0 \}$ and $S=\{ f:(-\infty,\infty) \to \mathbb{R} \vert f(d) < 0, \forall d < 0, f(d) > 0, \forall d > 0 \}$. From the class definition and
equ.~\ref{equ:worker_control}, an attraction force exists between all the robots
with distance $d_{ij} > d_{\rho}, \forall i,j \in E$ and a repulsive force 
acts on the robots with distance $d_{ij} < d_{\rho}$. The equilibrium point with
$u_i=0$ is when all the neighboring robots of robot $i$ are at distance
$d_{\rho}$ and with no guide robots within $d_{RFoV}$.

\subsection{Guide Swarm}
\label{sec:guide}
The guide swarm is the brains of the whole robot group, controlling the worker swarm, which is analogous to the muscles that do the work. 
Guides have the following high-level states: 1. Task Allocation, 2. Shaping setup, 3. Shaping, and 4. Movement. The high-level state machine of the guides is shown in fig.~\ref{fig:guide_state_machine}. Our depiction continues to assume a single shape set of shaping parameters. However, we can do multiple shapes in a single execution by activating dotted connections. All the state transitions between the four high-level states occur only with an agreement with all the guides. We use a barrier~\cite{rosbuzz2020}, a decentralized method that holds the state of the robots in a standby state until all the robots satisfy the conditions for the state transition.

\textbf{Task Allocation} The multi-robot task allocation
problem~\cite{korsah2013comprehensive} is a combinatorial optimization problem
requiring heuristics to obtain an approximate solution in polynomial time. In
our case, the Single Assignment (SA) problem and can be solved using auction or
consensus. The SA problem used by the guide robots to assign $t \in \Theta^s$ to
$\vert R^G \vert$ robots has the form:
\begin{align}
  \label{equ:SA-prob}
  &min \sum_{i \in  R^G } \sum_{j \in  \Theta^s } c_{ij}(x_i)a_{ij} \\
  &st. \sum_{j \in  \Theta^s} \leq 1, \forall i \in  R^G \text{ and }  \sum_{i \in R^G} \leq 1, \forall j \in  \Theta^{s} \nonumber
\end{align}
with $c_{ij} = \norm{x_i - x_j}$ being the cost function of assigning task $j$
to robot $i$ and $a_{ij} \in \{0,1\}$ being the binary assignment variable. We
solve the single assignment problem of equation~\ref{equ:SA-prob} using the
decentralized consensus described in~\cite{Varadharajan2020}.

\begin{figure*}
  \centering
  \includegraphics[width=0.325\linewidth]{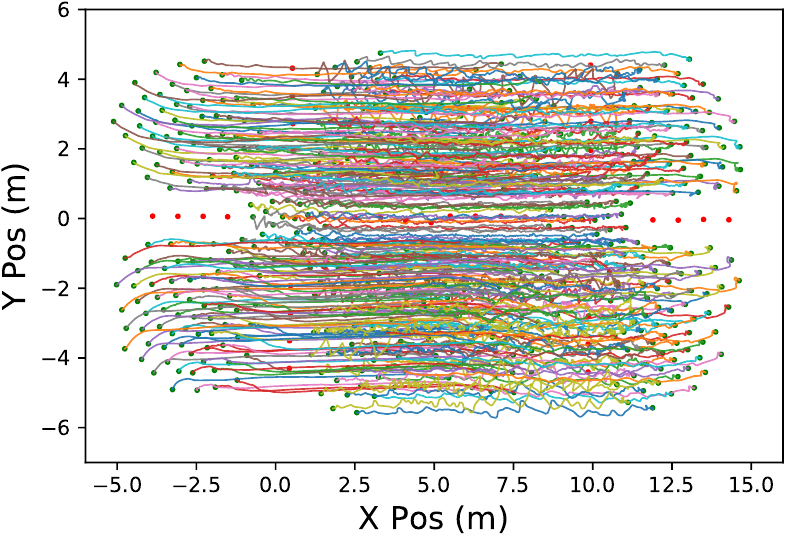}
  \includegraphics[width=0.325\linewidth]{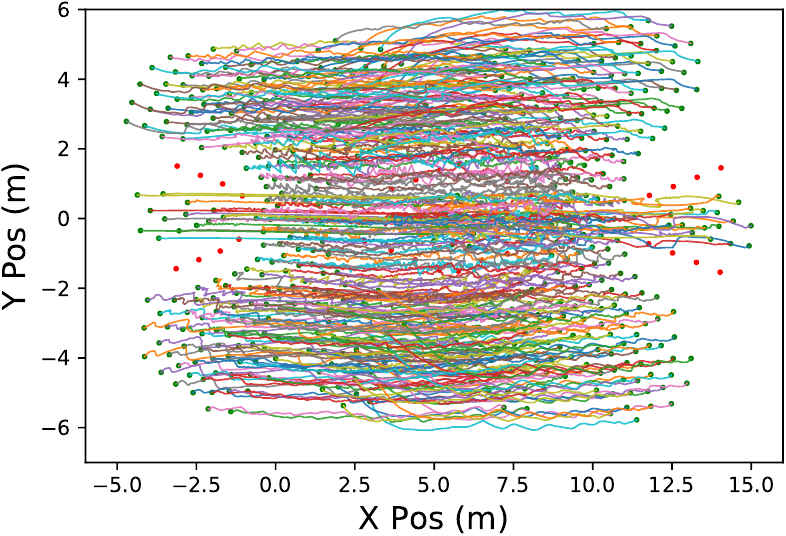}
  \includegraphics[width=0.325\linewidth]{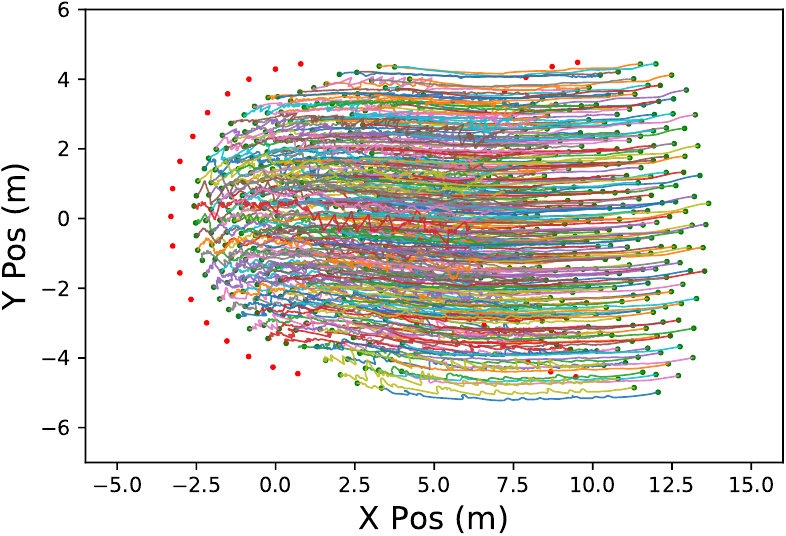}
  \caption{The different trajectories undertaken by robots in each shape: clover
    on the left, dumbbell in the center and circle on the right. Here we can
    already see the motion of the circle is more direct than the others, with
    the dumbbell showing noticeable vertical and horizontal distortion en route
    to its destination.}
  \label{fig:trajectory}
\end{figure*}

\begin{figure*}
  \centering
  \includegraphics[width=\linewidth]{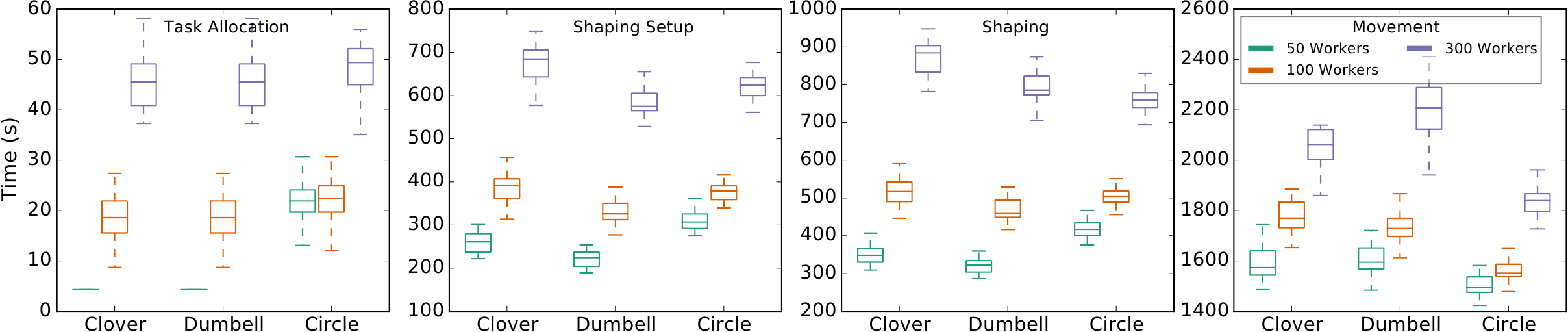}
  \caption{Time taken by the worker robots to transition from one state to
    another. We can see consistently longer and more varying times in the
    movement of clover and dumbbell shapes. The task allocation, shaping setup,
    and shaping of clovers and dumbbells also appear to be affected by
    increasing the swarm size.}
  \label{fig:sim_time}
\end{figure*}

\textbf{Shaping Setup and Shaping}
The guides edge-follow the worker swarm until they reach their assigned shape
angle. We refer the reader to the supplementary material in our code
repository~\cite{particleswarmrepo} for the details. The algorithm is partly
inspired from~\cite{Rubenstein795} and adds a recently-followed neighbor list to
escape concave edge areas (otherwise causing robots to be stuck on approach as
in~\cite{Rubenstein795}).
During edge-following, the robots maintain a list of ten recent neighbors to
avoid committing and following a previously followed neighbor. When a guide
comes in communication range with other guides, the robots coordinate to enable
simultaneous edge-following. In Shaping, the guide robots move into the worker
swarm at a given speed to attain the required shape.

\textbf{Movement} The task of translating in formation is defined as formation
shepherding, requiring the robots to preserve the edges $E$ in its communication
graph $\mathcal{G}$ (edge-consistent motion). This part of our solution is the
one mostly resembling the shepherding problem from literature.
The guide swarm uses this control law to translate towards the
goal while keeping the worker formation:
\begin{align}
  \label{equ:guides_control}
  u_{i} &= \alpha_g g_{i} + \beta_g c_{i} + \gamma_g a_{i} \\
    \label{equ:guides_control:contri_ang}
  a_{i} &= p(\theta_i^{tc}) \left( \perp \frac{x^c_i}{\norm{x^c_i}} \right) \\
   \label{equ:guides_control:contri_ang_piecewise}
  p(\theta) &=
  \begin{cases}
  -1 & ,  \theta < \pi \textit{ and } \theta < (\theta_{t} + \pi)  \textit{ and not}  (\theta < \theta_{t}) \\
   1 & ,  (\theta < \pi \textit{ and } \theta < (\theta_{t} + \pi)) \textit{ or } \theta > \theta_{t}  \\
  -1 & , otherwise
  \end{cases}
\end{align}

with $\theta_i^{tc} = \angle(x_i^c - x_t)$. The parameters $\alpha_g$, $\beta_g$
and $\gamma_g$ are design parameters respectively that define the importance of
moving towards the target ($g_{i}$), conserving the desired center of mass
distance to the workers ($c_{i}$) and maintaining the orientation between the
target and worker center of mass ($a_{i}$).
Both $g_{i}$ and $c_{i}$ use the harmonic function potential defined in
equ.~\ref{equ:harmonic_potential} as in equ.~\ref{equ:worker_control:contri1}.
Whereas the contribution to maintain the orientation is performed using
equ.~\ref{equ:guides_control:contri_ang}.

\section{Experiments}
We evaluate our approach both in simulation and on robots. All the code used in
the experiments is available as open source~\cite{particleswarmrepo}. We
implemented our approach using Buzz~\cite{pinciroli2016buzz} an extensible
domain-specific programming language for robot swarms. We used the Khepera IV
robot model in simulation and deployed customized Khepera IV robots equipped
with an Intel RealSense T265~\cite{realsense} and an Nvidia Jetson
TX1~\cite{nvidia_tx1} (some robots in the worker class were using a Raspberry Pi
4) as their on-board computer. Robots in simulation use a range and bearing
sensor (from our simulator, ARGoS~\cite{Pinciroli2012}) to communicate, while
the real robots use ad ad-hoc networking and local broadcast. We implemented
range and bearing in the real robots by creating data packets pairs with
[position, data] using the positional information computed from their T265
camera.
\begin{figure}
  \centering
  \includegraphics[width=0.99\linewidth]{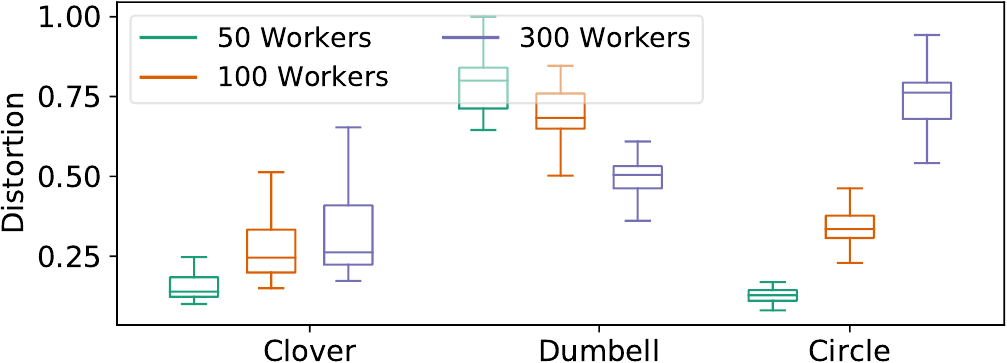}
  \caption{Distortion incurred by the swarm in different states computed using
    $\Upsilon$. Here we see how the dumbbell is particularly subject to
    distortion with smaller worker and guide swarms, although distortion
    decreases with larger worker swarms. On the other hand, the circle and
    clover shapes seem to have greater distortion with increasing worker swarm
    size.}
  \label{fig:sim_distortion}
\end{figure}

\subsection{Simulation}
Simulation experiments were performed using a physics-based
simulator, ARGoS~\cite{Pinciroli2012}. Simulation configuration used during our
experiments were: number of worker robots $N\in\{$50, 100, 300$\}$ and three
different shapes $Shape\in\{$Clover, Dumbbell, Circle$\}$ with 4 to 16
guides, 30 times per configuration with random initialization. We also run
experiments with 1000 robots and 32 guides for the clover shape.
All parameters used in the experiments can be found in our code
repository~\cite{particleswarmrepo}. Fig.~\ref{fig:shapes} shows the
configurations, and Fig.~\ref{fig:trajectory} the trajectories of the robots
moving the shapes 10m from the initial location.

\begin{figure}
  \centering
  \includegraphics[width=\linewidth]{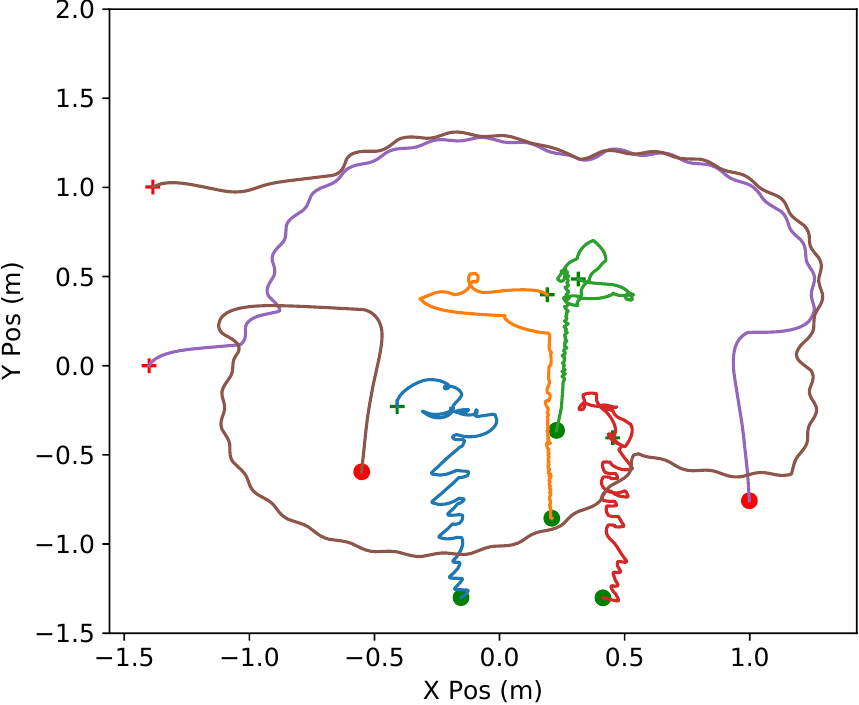}
  \includegraphics[width=\linewidth]{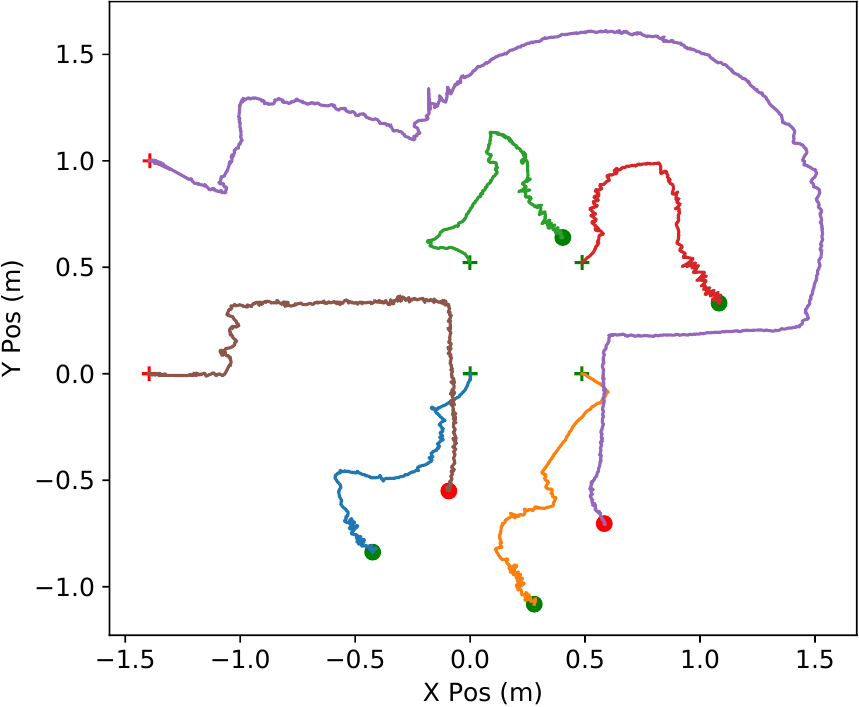}
  \includegraphics[width=\linewidth]{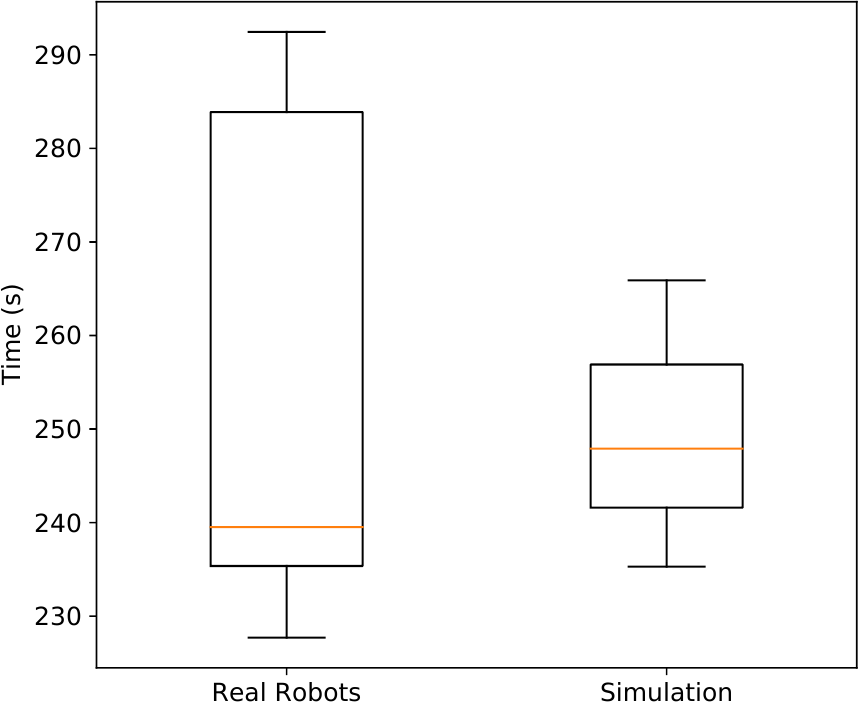}
  \caption{Trajectory for shaping and translating a swarm of 6 real robots.
    top: simulation; middle: real robots; bottom: the time taken for the
    operation. Slightly different actuation and forces affecting
    the real robot dynamics make for a smoother set of trajectories
    simulation, with minor drift.}
  \label{fig:real}
\end{figure}


Our primary metric for performance is the distortion $\Upsilon$:
$ \Upsilon = \frac{1}{N} \sum_{i \in R_W} \sum_{j \in R_W, j \neq i } \norm{ x_ij(t_s)} - \norm{x_ij(t_f)}
$
with $t_s$ being the time at shaping and $t_f$ being the time at reaching the
target, which quantifies the deformation of the shape after motion. $\Upsilon$
is normalized in $[0,1]$ to the largest value from all the configurations to
make comparison possible. A value of 0 is a perfect displacement with no
distortion (i.e., no relative change in distances between the robots), while 1
is the highest distortion seen in the current formation. The distortion metric
can be also viewed as the relative change in position of a robot with respect to
the worst case encountered in the experimental configuration.

Fig.~\ref{fig:sim_time} breaks down the robot activities as described in Section~\ref{sec:guide}: \emph{Task
  Allocation} for the guide robots to select shaping targets, \emph{Shaping
  Setup} to locate the worker swarm and edge-follow to enclose it,
\emph{Shaping} to create a desired formation with the worker, and
\emph{Movement} takes the desired formation to a target. \emph{Task Allocation}
has a monotonic increase in time with number of robots and does not vary
with configurations since it is a function of number of robots and available
tasks.
\emph{Shaping Setup} and \emph{Shaping} exhibit a similar pattern that depends
on the configuration: this is due to the fact that both require reaching a given
alignment.
\emph{Movement} exhibits different dynamics that depend on each shape and number
of workers.
It is worth noting that the circle shape took the least amount of time because
the robots require fewer corrective moves to maintain formation and spend more
time moving towards the target.

Fig.~\ref{fig:sim_distortion} compares the distortion from motion for various
shapes. 
For Clover and Circle, the distortion increases with the number of robots, but
decreases for Dumbbell, although it is generally higher for the latter. We
believe that the higher number of guides mitigate the distortion for the more
complex shape of the Dumbbell.


\subsection{Physical robots}
We used 6 Khepera IV robots with 4 as workers and 2 as guides.
We used a simple square like shape to be formed by the robots. The robots were
able to task allocate, preform the shaping setup to reach the provided shaping
targets and move to a target of about 1m in formation. Fig.~\ref{fig:real} top
shows the trajectory of the robots for a similar shape in simulation, middle,
the trajectory of the real robots and the bottom, shows the time taken to reach
the target in formation. The experiments reported 240s on average with real
robots and 247s in simulation, matching expectations.

\section{Conclusions} 
\label{sec:conclusion}

We propose a hierarchical approach to control a swarm of robots that form shapes
and translate in formation with two types of robots: a large group of simple
robots (workers) and a small swarm of intelligent robots (guides). The process
of decision-making for formation and translation to targets is performed by the
guides, and workers act as the muscle in forming shapes with no knowledge of the
overall task. The interactions of the swarms are designed to be simple and local
through virtual potential fields that depend on the relative position of the
robots. This approach relies on the knowledge of the guide swarm, which
coordinates to obtain the desired shape of the worker swarm, like smart
particles. The interaction properties of the worker swarm are dynamically
adapted by the guide swarm, leveraging its additional sensing and knowledge. 
The prototype framework implemented in this work lays the foundation for
hierarchically self-organizing swarms. Potential uses of the approach could be
targeted drug delivery, cancer treatment, and smart materials.

\bibliographystyle{IEEEtran}
\bibliography{references}

\begin{thebibliography}{10}
\providecommand{\url}[1]{#1}
\csname url@rmstyle\endcsname
\providecommand{\newblock}{\relax}
\providecommand{\bibinfo}[2]{#2}
\providecommand\BIBentrySTDinterwordspacing{\spaceskip=0pt\relax}
\providecommand\BIBentryALTinterwordstretchfactor{4}
\providecommand\BIBentryALTinterwordspacing{\spaceskip=\fontdimen2\font plus
\BIBentryALTinterwordstretchfactor\fontdimen3\font minus
  \fontdimen4\font\relax}
\providecommand\BIBforeignlanguage[2]{{%
\expandafter\ifx\csname l@#1\endcsname\relax
\typeout{** WARNING: IEEEtran.bst: No hyphenation pattern has been}%
\typeout{** loaded for the language `#1'. Using the pattern for}%
\typeout{** the default language instead.}%
\else
\language=\csname l@#1\endcsname
\fi
#2}}

\bibitem{Dorigo2021}
M.~Dorigo, G.~Theraulaz, and V.~Trianni, ``{Swarm robotics: Past, present, and
  future},'' \emph{Proceedings of the IEEE}, vol. 109, no.~7, pp. 1152--1165,
  2021.

\bibitem{Dorigo2020}
------, ``{Reflections on the future of swarm robotics},'' \emph{Science
  Robotics}, vol.~5, no.~49, pp. 9--12, 2020.

\bibitem{scheutz2007real}
M.~Scheutz, ``Real-time hierarchical swarms for rapid adaptive multi-level
  pattern detection and tracking,'' in \emph{2007 IEEE Swarm Intelligence
  Symposium}.\hskip 1em plus 0.5em minus 0.4em\relax IEEE, 2007, pp. 234--241.

\bibitem{chen2010hierarchical}
H.~Chen, Y.~Zhu, K.~Hu, and X.~He, ``Hierarchical swarm model: a new approach
  to optimization,'' \emph{Discrete Dynamics in Nature and Society}, vol. 2010,
  2010.

\bibitem{Zhang2023}
Y.~Zhang, S.~Oğuz, S.~Wang, E.~Garone, X.~Wang, M.~Dorigo, and M.~K. Heinrich,
  ``Self-reconfigurable hierarchical frameworks for formation control of robot
  swarms,'' \emph{IEEE Transactions on Cybernetics}, pp. 1--14, 2023.

\bibitem{hu2020occlusion}
J.~Hu, A.~E. Turgut, T.~Krajn{\'\i}k, B.~Lennox, and F.~Arvin,
  ``Occlusion-based coordination protocol design for autonomous robotic
  shepherding tasks,'' \emph{IEEE Transactions on Cognitive and Developmental
  Systems}, 2020.

\bibitem{razali2013flock}
S.~Razali, N.~F. Shamsudin, M.~Osman, Q.~Meng, and S.-H. Yang, ``Flock
  identification using connected components labeling for multi-robot
  shepherding,'' in \emph{2013 International Conference on Soft Computing and
  Pattern Recognition (SoCPaR)}.\hskip 1em plus 0.5em minus 0.4em\relax IEEE,
  2013, pp. 298--303.

\bibitem{ozdemir2017shepherding}
A.~{\"O}zdemir, M.~Gauci, and R.~Gro{\ss}, ``Shepherding with robots that do
  not compute,'' in \emph{ECAL 2017, the Fourteenth European Conference on
  Artificial Life}.\hskip 1em plus 0.5em minus 0.4em\relax MIT Press, 2017, pp.
  332--339.

\bibitem{huang2015multifunctional}
C.~Huang, G.~Yang, Q.~Ha, J.~Meng, and S.~Wang, ``Multifunctional “smart”
  particles engineered from live immunocytes: toward capture and release of
  cancer cells,'' \emph{Advanced Materials}, vol.~27, no.~2, pp. 310--313,
  2015.

\bibitem{particleswarmrepo}
``Particle swarm code repository,''
  \url{https://github.com/MISTLab/Particle_swarm.git}, 2023.

\bibitem{Slavkov2018}
I.~Slavkov, D.~Carrillo-Zapata, N.~Carranza, X.~Diego, F.~Jansson, J.~A.
  Kaandorp, S.~Hauert, and J.~Sharpe, ``{Morphogenesis in Robot Swarms},''
  \emph{Accepted for publication on Science Robotics. Available on Dec 19,
  2018}, vol.~3, no.~24, pp. 1--17, 2018.

\bibitem{Rubenstein795}
M.~Rubenstein, A.~Cornejo, and R.~Nagpal, ``{Programmable self-assembly in a
  thousand-robot swarm},'' \emph{Science}, vol. 345, no. 6198, pp. 795--799,
  2014.

\bibitem{Vasarhelyi2018}
G.~V{\'{a}}s{\'{a}}rhelyi, C.~Vir{\'{a}}gh, G.~Somorjai, T.~Nepusz, A.~E.
  Eiben, and T.~Vicsek, ``{Optimized flocking of autonomous drones in confined
  environments},'' \emph{Science Robotics}, vol.~3, no.~20, pp. 1--14, 2018.

\bibitem{ramos2019evolving}
R.~P. Ramos, S.~M. Oliveira, S.~M. Vieira, and A.~L. Christensen, ``Evolving
  flocking in embodied agents based on local and global application of
  reynolds’ rules,'' \emph{Plos one}, vol.~14, no.~10, p. e0224376, 2019.

\bibitem{fine2013unifying}
B.~T. Fine and D.~A. Shell, ``Unifying microscopic flocking motion models for
  virtual, robotic, and biological flock members,'' \emph{Autonomous Robots},
  vol.~35, no.~2, pp. 195--219, 2013.

\bibitem{reynolds1987flocks}
C.~W. Reynolds, ``Flocks, herds and schools: A distributed behavioral model,''
  in \emph{Proceedings of the 14th annual conference on Computer graphics and
  interactive techniques}, 1987, pp. 25--34.

\bibitem{dimarogonas2006leader}
D.~V. Dimarogonas, M.~Egerstedt, and K.~J. Kyriakopoulos, ``A leader-based
  containment control strategy for multiple unicycles,'' in \emph{Proceedings
  of the 45th IEEE Conference on Decision and Control}.\hskip 1em plus 0.5em
  minus 0.4em\relax IEEE, 2006, pp. 5968--5973.

\bibitem{ji2006leader}
M.~Ji, A.~Muhammad, and M.~Egerstedt, ``Leader-based multi-agent coordination:
  Controllability and optimal control,'' in \emph{2006 American Control
  Conference}.\hskip 1em plus 0.5em minus 0.4em\relax IEEE, 2006, pp. 6--pp.

\bibitem{panagou2014cooperative}
D.~Panagou and V.~Kumar, ``Cooperative visibility maintenance for
  leader--follower formations in obstacle environments,'' \emph{IEEE
  Transactions on Robotics}, vol.~30, no.~4, pp. 831--844, 2014.

\bibitem{bjorkenstam2006leader}
S.~Bj{\"o}rkenstam, M.~Ji, M.~B. Egerstedt, and C.~F. Martin, ``Leader-based
  multi-agent coordination through hybrid optimal control.''\hskip 1em plus
  0.5em minus 0.4em\relax Georgia Institute of Technology, 2006.

\bibitem{wiech2018virtual}
J.~Wiech, V.~A. Eremeyev, and I.~Giorgio, ``Virtual spring damper method for
  nonholonomic robotic swarm self-organization and leader following,'' 2018.

\bibitem{hoshi2018robustness}
H.~Hoshi, I.~Iimura, S.~Nakayama, Y.~Moriyama, and K.~Ishibashi, ``Robustness
  of herding algorithm with a single shepherd regarding agents' moving
  speeds,'' \emph{Journal of Signal Processing}, vol.~22, no.~6, pp. 327--335,
  2018.

\bibitem{chaimowicz2007aerial}
L.~Chaimowicz and V.~Kumar, ``Aerial shepherds: Coordination among uavs and
  swarms of robots,'' in \emph{Distributed Autonomous Robotic Systems 6}.\hskip
  1em plus 0.5em minus 0.4em\relax Springer, 2007, pp. 243--252.

\bibitem{long2021shepherding}
N.~K. Long, M.~Garratt, K.~Sammut, D.~Sgarioto, and H.~A. Abbass, ``Shepherding
  autonomous goal-focused swarms in unknown environments using hilbert
  space-filling paths,'' \emph{Shepherding UxVs for Human-Swarm Teaming: An
  Artificial Intelligence Approach to Unmanned X Vehicles}, pp. 31--50, 2021.

\bibitem{kubo2022herd}
M.~Kubo, M.~Tashiro, H.~Sato, and A.~Yamaguchi, ``Herd guidance by multiple
  sheepdog agents with repulsive force,'' \emph{Artificial Life and Robotics},
  pp. 1--12, 2022.

\bibitem{Varadharajan2020}
V.~S. Varadharajan, D.~St-Onge, B.~Adams, and G.~Beltrame, ``Swarm relays:
  Distributed self-healing ground-and-air connectivity chains,'' \emph{IEEE
  Robotics and Automation Letters}, vol.~5, no.~4, pp. 5347--5354, 2020.

\bibitem{rosbuzz2020}
D.~St-Onge, V.~S. Varadharajan, I.~{\v{S}}vogor, and G.~Beltrame, ``From design
  to deployment: decentralized coordination of heterogeneous robotic teams,''
  \emph{Frontiers in Robotics and AI}, vol.~7, p.~51, 2020.

\bibitem{pinciroli2016tuple}
C.~Pinciroli, A.~Lee-Brown, and G.~Beltrame, ``A tuple space for data sharing
  in robot swarms,'' in \emph{Proceedings of the 9th EAI International
  Conference on Bio-inspired Information and Communications Technologies
  (formerly BIONETICS)}, 2016, pp. 287--294.

\bibitem{Zhang2011}
H.~T. Zhang, C.~Zhai, and Z.~Chen, ``{A general alignment repulsion algorithm
  for flocking of multi-agent systems},'' \emph{IEEE Transactions on Automatic
  Control}, vol.~56, no.~2, pp. 430--435, 2011.

\bibitem{stuart2000reactive}
S.~J. Stuart, A.~B. Tutein, and J.~A. Harrison, ``A reactive potential for
  hydrocarbons with intermolecular interactions,'' \emph{The Journal of
  chemical physics}, vol. 112, no.~14, pp. 6472--6486, 2000.

\bibitem{korsah2013comprehensive}
G.~A. Korsah, A.~Stentz, and M.~B. Dias, ``A comprehensive taxonomy for
  multi-robot task allocation,'' \emph{The International Journal of Robotics
  Research}, vol.~32, no.~12, pp. 1495--1512, 2013.

\bibitem{pinciroli2016buzz}
C.~Pinciroli and G.~Beltrame, ``Buzz: An extensible programming language for
  heterogeneous swarm robotics,'' in \emph{2016 IEEE/RSJ International
  Conference on Intelligent Robots and Systems (IROS)}.\hskip 1em plus 0.5em
  minus 0.4em\relax IEEE, 2016, pp. 3794--3800.

\bibitem{realsense}
``{Intel Realsense Tracking Camera},''
  \url{https://www.intelrealsense.com/tracking-camera-t265/}, 2023, [Online;
  accessed 29-July-2023].

\bibitem{nvidia_tx1}
``{NVIDIA TX1 product website},''
  \url{https://developer.nvidia.com/embedded/jetson-tx1}, 2023, [Online;
  accessed 29-July-2023].

\bibitem{Pinciroli2012}
C.~Pinciroli, V.~Trianni, R.~O’Grady, G.~Pini, A.~Brutschy, M.~Brambilla,
  N.~Mathews, E.~Ferrante, G.~Di~Caro, F.~Ducatelle, \emph{et~al.}, ``Argos: a
  modular, parallel, multi-engine simulator for multi-robot systems,''
  \emph{Swarm intelligence}, vol.~6, no.~4, pp. 271--295, 2012.

\end{thebibliography}

\end{document}